\documentclass{article}

\usepackage[preprint]{neurips_2026}

\usepackage[utf8]{inputenc} 
\usepackage[T1]{fontenc}    
\usepackage{hyperref}       
\usepackage{url}            
\usepackage{booktabs}       
\usepackage{amsfonts}       
\usepackage{nicefrac}       
\usepackage{microtype}      
\usepackage{xcolor}         
\usepackage{amsmath}
\usepackage{graphicx}
 \usepackage{multirow}
 \usepackage{pifont}
 \usepackage{standalone}
\usepackage[most]{tcolorbox}

\title{AnchorEdit: Maintaining Temporal Consistency in Multi-turn Image Editing via Causal Memory}

\author{Hang Xu\textsuperscript{1},
Xiaoxiao Ma\textsuperscript{1},  
Guohui Zhang\textsuperscript{1}, 
Hu Yu\textsuperscript{1},
Siming Fu\textsuperscript{2},  
Jie Huang\textsuperscript{2}, \\
\textbf{Lin Song}\textsuperscript{2}, 
\textbf{Haoyang Huang}\textsuperscript{2}, 
\textbf{Nan Duan}\textsuperscript{2}, 
\textbf{Feng Zhao}\textsuperscript{1\dag} \\
\textsuperscript{1}University of Science and Technology of China,
\textsuperscript{2}JD Explore Academy,
}

\begin{document}

{
  \renewcommand{\thefootnote}{\fnsymbol{footnote}}
  \footnotetext{\textsuperscript{\dag}Corresponding author.}
}

\maketitle

\begin{abstract}

Multi-turn image editing is essential for iterative design, yet current models often struggle with identity drift and error accumulation over successive steps. While existing research leverages video priors for consistency, their reliance on bidirectional attention is fundamentally misaligned with the causal, sequential nature of interactive editing. In this paper, we propose AnchorEdit, the first autoregressive (AR) diffusion-based framework designed specifically for high-resolution, long-term multi-turn editing. AnchorEdit bridges the gap between video priors and causal inference through a three-stage training curriculum: identity-preserving sing-turn pretraining, causal AR forcing fine-tuning with a novel self-rollout strategy to mitigate exposure bias, and consistency distillation for efficient 4-step generation. During inference, we introduce a memory mechanism to anchor the initial subject identity and ensure stable extrapolation across extended editing trajectories. To evaluate performance, we provide a new high-resolution multi-turn editing benchmark designed to stress-test long-horizon stability. Extensive experiments demonstrate that AnchorEdit achieves state-of-the-art results, maintaining exceptional subject fidelity and instruction following even over 10+ interaction rounds.
\end{abstract}
\section{Introduction}
\label{sec:intro}
Natural-language image editing is now essential for visual content creation~\cite{brooks2023instructpix2pix,kawar2023imagic,zhang2024magicbrush,sheynin2023emuedit,yu2025anyedit}. Yet, practical workflows are typically iterative, requiring users to refine images through a sequence of modifications to objects, styles, and details. While current models excel at single-turn tasks~\cite{wu2025qwen,team2026firered}, they often struggle with multi-turn editing, failing to preserve subject identity and visual coherence as errors from earlier steps accumulate over successive turns.


Due to the scarcity of multi-turn editing data, recent research leverages video data and pre-trained video generation models~\cite{wan2025wan,yang2025cogvideox,kong2025hunyuanvideo} for supervision. These models offer rich priors on object permanence and identity preservation, which are essential for maintaining consistency across sequential edits, as illustrated in Fig. \ref{fig:first}(a). Recent methods such as ChronoEdit~\cite{wu2025chronoedit}, CoF-T2I~\cite{tong2026cof}, and VINCIE~\cite{qu2025vincie} further validate the effectiveness of using video-based priors for consistent image editing.


However, existing video-based editing methods are largely built on bidirectional or full-context generation, which is fundamentally misaligned with the causal structure of interactive multi-turn editing as shown in Fig. \ref{fig:first}(b). In practice, each edit should depend only on the source image and previous states, whereas earlier results cannot incorporate future instructions. This discrepancy often results in degraded instruction following, identity drift, and accumulated visual inconsistencies during deployment. Furthermore, the use of fixed-length contexts in these models restricts their ability to extrapolate to longer editing chains and prevents truly online interaction.


These observations suggest two key desiderata for long-horizon multi-turn editing: \textit{leveraging the rich visual priors of video generation model} and \textit{aligning with the causal nature of sequential editing}. Autoregressive (AR) video generation~\cite{huang2025selfforcing,yin2025causvid,zhu2026causal,liu2025rollingforcing,yang2025longlive,chen2025skyreelsv2} provides a natural framework for this, as it models sequences causally and supports extrapolation beyond fixed horizons.
However, directly applying AR video generation to editing is challenging. Unlike standard video continuation, multi-turn editing requires precise, instruction-driven state transitions and strict identity preservation. Because each generated edit serves as the condition for subsequent turns, minor artifacts or identity deviations can be recursively amplified. Therefore, an effective causal editor must incorporate specific mechanisms for identity anchoring and robustness to error accumulation in self-generated contexts.

In this work, we introduce AnchorEdit, a comprehensive framework for high-resolution, long-term multi-turn image editing. The model development follows a systematic three-stage training regime. Initially, we fine-tune a video model for identity preservation, using "no-change" reconstruction tasks and increased RoPE~\cite{heo2024rotary,su2023roformer,ding2024longrope,peng2026yarn} distance between frames to lock in subject fidelity. Building upon this, the second stage employs causal frame-autoregressive training. We implement a self-rollout~\cite{huang2025self} strategy—randomly replacing ground-truth inputs with model predictions—and apply progressive loss weighting to later turns to mitigate exposure bias and rectify cumulative errors. Finally, we use adversarial consistency distillation~\cite{yin2024improved} to compress the model into an efficient 4-step generator.
During inference, we propose a specialized memory mechanism and a manifold-constrained RoPE strategy to enable robust extrapolation. By anchoring the initial frame as a global reference and maintaining a historical sliding window~\cite{xiao2024streamllm}, we ensure the model attends to consistent features. Crucially, we keep RoPE indices fixed within the trained range regardless of sequence length. This prevents the attention mechanism from entering untrained manifolds, ensuring stable, high-quality results across ten or more interaction rounds.

To address the lack of standardized evaluation tools, we introduce a comprehensive multi-turn editing benchmark featuring high-resolution (1024p) data and diverse sequence depths, specifically designed to stress-test model stability across extended interactions. Extensive experiments demonstrate that AnchorEdit not only maintains state-of-the-art performance in single-turn tasks but also significantly surpasses existing methods in maintaining long-term subject identity and visual quality. Our primary contributions are summarized as follows:

\begin{itemize}
\item \textbf{First High-Resolution AR Diffusion Framework for Multi-turn Image Editing}: We present AnchorEdit, the first autoregressive diffusion-based model designed specifically for high-resolution, long-term multi-turn image editing. 
\item \textbf{Robust Forcing Training and Memory Mechanisms}: We are the first to introduce self-rollout forcing and historical memory mechanisms into the multi-turn editing paradigm. These strategies effectively mitigate exposure bias and cumulative errors, ensuring strict identity consistency and stable subject preservation over ten or more editing rounds.
\item \textbf{Comprehensive Benchmark and State-of-the-Art Performance}: We provide a new high-resolution multi-turn editing benchmark and a scalable data pipeline to support the community. Extensive experiments demonstrate that AnchorEdit achieves state-of-the-art results in both instruction adherence and visual coherence compared to existing methods.
\end{itemize}

\section{Related work}
\label{sec:related}

\subsection{Image Editing from Video Data and Video Generation Models}
To enhance subject consistency, contemporary models increasingly leverage video data for its inherent temporal coherence, high realism, and cost-effective abundance compared to specialized image datasets. Recent research favors video-based frameworks over image-only ones, often incorporating "inference frames" to bridge the visual gap between original and edited states (e.g., Frame2Frame~\cite{rotstein2025pathways}, ChronoEdit~\cite{wu2025chronoedit}, COF-T2I~\cite{tong2026cof}). Furthermore, works like VINCIE~\cite{qu2025vincie} and ChronoEdit demonstrate that complex editing logic and multi-turn capabilities can be robustly acquired directly from video sequences, providing a foundational source for training high-fidelity editing tasks.

\subsection{Autoregressive Video Diffusion Models}
While bidirectional video models are widely used, they are often limited by fixed temporal horizons and high computational costs. Autoregressive (AR) video diffusion models~\cite{yin2025slow,zhu2026causal,liu2025rolling,cui2025self,yuan2026helios} overcome these constraints by generating frames sequentially, offering the flexibility to introduce new prompts for each turn—a key advantage for multi-turn editing. Predominant training paradigms include Teacher Forcing~\cite{zhou2025taming}, which conditions generation on clean history, and Diffusion Forcing~\cite{chen2024diffusion}, which uses noisy frames to mitigate the training-inference discrepancy. Recent methods further improve AR video generation through self-generated rollouts and causal distillation~\cite{huang2025self,zhu2026causal}, as well as long-context memory and RoPE-based temporal extrapolation~\cite{chen2026contextforcing,xiao2025knotforcing,wei2025videorope,yesiltepe2026infinityrope}.

\subsection{Multi-turn Image Editing}
Multi-turn editing requires global consistency across iterative instructions. Early methods relied on training-free optimization, such as DDIM inversion (e.g., Multi-turn Consistent Editing~\cite{zhou2025multi}, etc~\cite{hertz2022prompt2prompt,mokady2022nulltext,tumanyan2022plugandplay,cao2023masactrl}.), but often struggled with limited generalization. Recent research has shifted toward leveraging video generation models to provide consistency priors~\cite{qu2025vincie}. However, despite their improved robustness, current approaches still face critical bottlenecks in inference speed, output resolution, and temporal extrapolation, preventing them from becoming practical, application-level solutions.

\section{AnchorEdit: Enhancing multi-turn consistency with forcing and memory}

\subsection{Preliminary: Autoregressive Video Generation and Multi-turn Image Editing}

\begin{figure*}[t]
\centering
\includegraphics[width=0.95\linewidth]{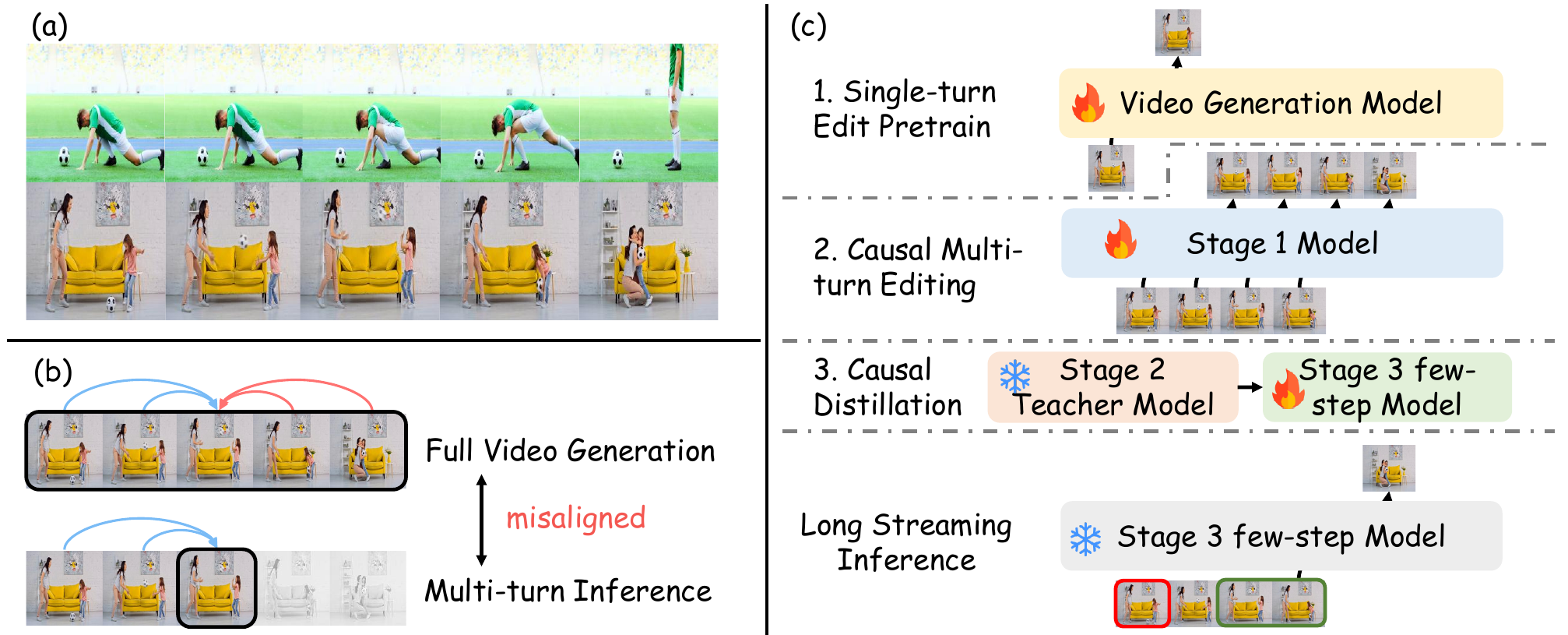}
\caption{\textbf{Motivation and Framework Overview.} \textbf{Top-left}: Video sequences provide natural examples of continuous subject evolution, serving as a rich data source for multi-turn editing. \textbf{Bottom-left}: Traditional video generation models rely on bidirectional attention, which is misaligned with the sequential, causal nature of multi-turn inference. \textbf{Right}: To bridge this gap, we propose a progressive three-stage training and inference pipeline—comprising single-turn pretraining, causal multi-turn optimization, and distillation. This results in an autoregressive (AR) diffusion model capable of maintaining exceptional long-term consistency across extended editing trajectories.}
\label{fig:first}
\end{figure*}

Autoregressive (AR) video generation models a video as a sequence of visual states generated in causal order. Given a text prompt $c$ and a video sequence $\{x_i\}_{i=1}^{N}$, the joint distribution is factorized as:
\begin{equation}
p_\theta(x_{1:N}\mid c) = \prod_{i=1}^{N} p_\theta(x_i \mid x_{<i}, c)
\end{equation}
where $x_{<i}=\{x_1,\ldots,x_{i-1}\}$ denotes the generated history. This causal procedure supports streaming generation and length extrapolation, as each new visual state is produced incrementally by conditioning on previous context without access to future information.
Multi-turn image editing can be viewed as an instruction-conditioned extension of this process. Given a source image $x_0$ and a sequence of editing instructions $\{c_i\}_{i=1}^{N}$, each edit corresponds to a visual state, and each instruction acts as a turn-wise condition driving the transition. By replacing the global video prompt $c$ with turn-specific instructions $c_i$, and treating the source image $x_0$ as a persistent visual anchor, the AR objective naturally adapts into a causal formulation for multi-turn editing:
\begin{equation}
p_\theta(x_i \mid x_{<i}, c) \quad \Rightarrow \quad p_\theta(x_i \mid x_0, x_{<i}, c_i)
\end{equation}
This formulation aligns with the logic of interactive editing: the $i$-th edit depends strictly on the source image, the history of edited states, and the current instruction.
This correspondence motivates us to build AnchorEdit on a video generation backbone, treating iterative modifications as a continuous temporal trajectory. By framing each editing turn as a state transition within a sequence, we leverage the robust temporal and structural priors inherent in video models. To bridge the gap between video continuation and instruction-guided editing which requires strict identity preservation and robustness to self-generated histories, AnchorEdit employs a staged causal training regime and anchor-based inference, as detailed in the following sections.

\subsection{Multi-stage Causal Training}

As shown in Fig. \ref{fig:first}(c), Starting from a pre-trained video generation backbone, we implement a three-stage progressive training process: (1) single-turn editing pretraining, (2) causal multi-turn adaptation, and (3) few-step distillation. This progression yields a high-performance editing model that balances superior visual quality with rapid inference speed.

\begin{figure*}[t]
\centering
\includegraphics[width=0.95\linewidth]{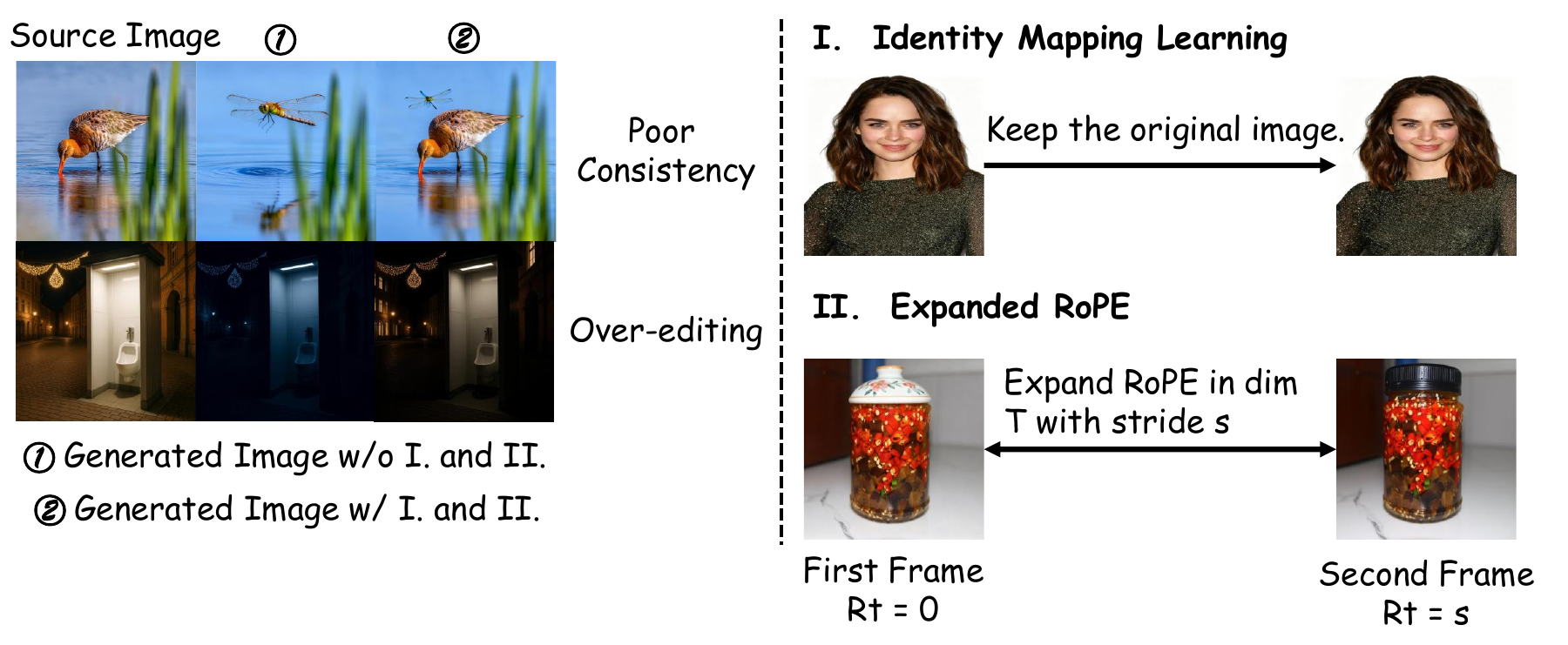}
\caption{\textbf{Stage 1 training strategies for enhanced editing consistency.} \textbf{Left:} Visual comparison. Without our proposed strategies (\ding{172}), the model suffers from poor consistency and over-editing. Our approach (\ding{173}) effectively preserves identity and structural integrity. \textbf{Right:} Core components: (I) \textit{Identity Mapping Learning} to enforce source reconstruction; (II) \textit{Expanded RoPE}, which increases the relative temporal distance between the source and edited states using a stride $s$ to prevent semantic collapse during discrete transitions.}
\label{fig:phase1}
\end{figure*}

\paragraph{Stage 1: Single-turn Editing Pretraining.}

The first stage adapts the video generation backbone to instruction-guided image editing by representing each edit as a two-state sequence. The source and edited images are encoded as latents $\mathbf{z}_1, \mathbf{z}_2 \in \mathbb{R}^{C \times H \times W}$ and concatenated along the temporal dimension, allowing the backbone to model the edit as a transition from source to target, as shown in Fig. \ref{fig:phase1}.

\emph{Expanding RoPE distance for discrete edit transitions(I in Fig. \ref{fig:phase1})}
Standard video models assume small relative RoPE distances between frames to encourage smooth motion. However, editing often requires discrete semantic changes. We therefore enlarge the relative RoPE distance between states from $\Delta=1$ to $\Delta=S$ ($S>1$). This weakens the backbone's smooth-motion bias and provides more capacity for localized semantic modifications without altering unrelated regions.

\emph{Null-instruction training for identity preservation(II in Fig. \ref{fig:phase1}).}
To mitigate identity drift, we augment the training with null-instruction reconstruction. Given a source latent $\mathbf{z}_{\mathrm{src}}$ and a null instruction $c_{\mathrm{null}}$, the model is trained to minimize:
\begin{equation}
    \mathcal{L}_{\mathrm{id}} = \left\| \mathcal{F}_{\theta}(\mathbf{z}_{\mathrm{src}}, c_{\mathrm{null}}) - \mathbf{z}_{\mathrm{src}} \right\|_2^2
\end{equation}
This objective regularizes the model toward an identity mapping, encouraging the preservation of regions unrelated to the editing instruction.

Overall, this stage converts the generative backbone into an instruction-guided editor with minimal perturbation to the original distribution. Unlike video-based editing methods that rely on auxiliary latent frames to bridge the source and target, AnchorEdit uses only the source and edited states, significantly reducing memory overhead for high-resolution tasks.


=\% Temporally-Progressive Objective:
To force the model to handle increasing error accumulation, we design a curriculum where both the task difficulty and supervision intensity increase with the frame index $i$. Specifically, we set:


\paragraph{Stage 2: Multi-turn Causal Autoregressive Training.} Given that multi-turn editing is inherently a sequential process, we convert the generation process to a causal autoregressive formulation by applying causal attention mask in Fig. \ref{fig:phase2}. We adopt Diffusion Forcing to maintain efficiency, where the model learns to predict the $i$-th clean frame $\mathbf{z}_{0,i}$ conditioned on previous noisy frames $\mathbf{z}_{t,<i}$. The central challenge is the exposure bias, where cumulative errors in $\mathbf{z}_{0,<i}$ lead to divergence. To mitigate this, we implement three synergistic strategies:

\begin{figure*}[t]
\centering
\includegraphics[width=0.95\linewidth]{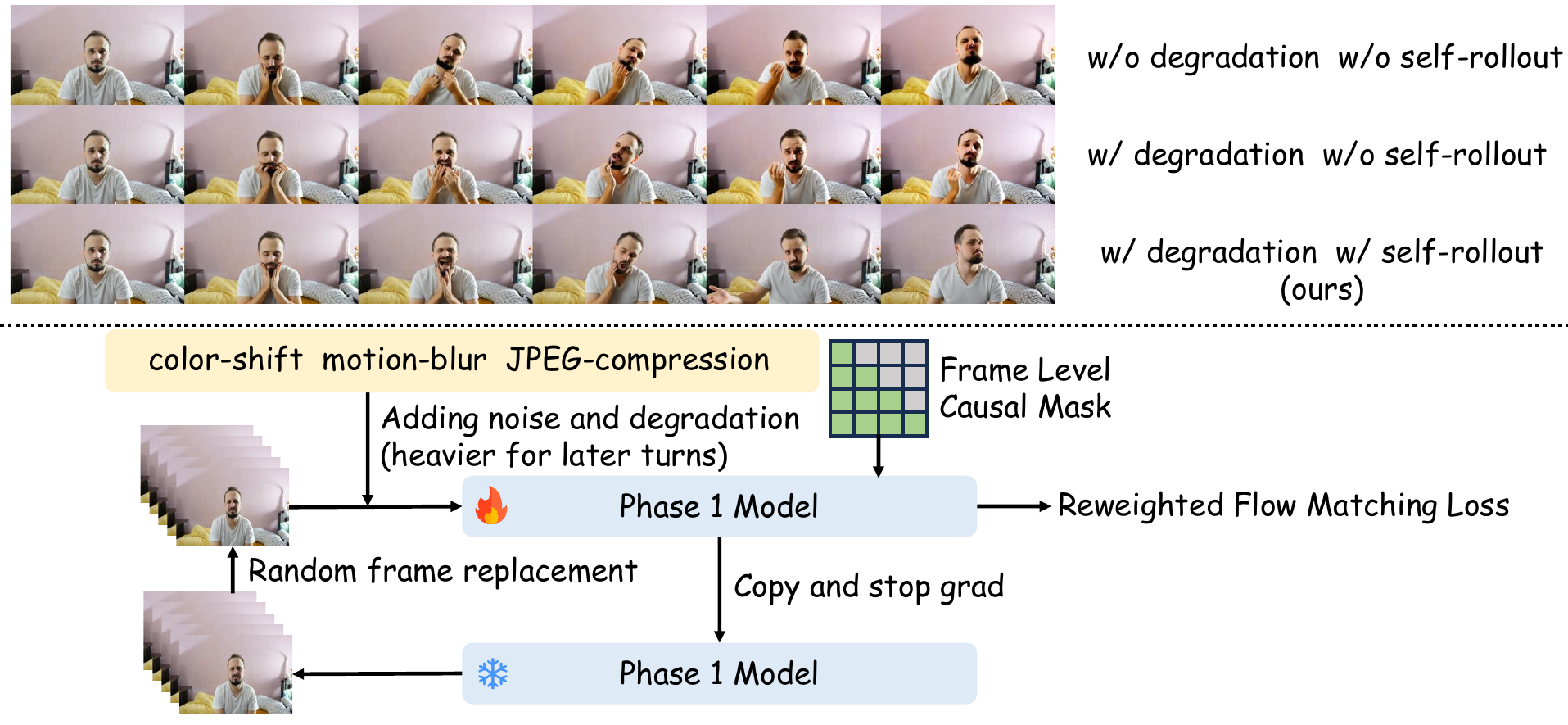}
\caption{\textbf{Stage 2: Multi-turn Consistency Training and Ablation.}
\textbf{Top (Ablation):} Visual comparison of training configurations. Without degradation injection (Row 1), the model accumulates artifacts. Without self-rollout (Row 2), distribution drift occurs. Our full design (Row 3) ensures long-term stability.
\textbf{Bottom (Methodology):} The training pipeline uses a frame-level causal mask to model sequential dependencies. To enhance robustness, we incorporate synthetic degradation and self-rollout frames. A temporally-progressive strategy scales degradation intensity, noise levels, and loss weights per turn to effectively regularize multi-turn consistency.}
\label{fig:phase2}
\end{figure*}

\textit{Supervised training with synthetic degradation injection.} We simulate inference-time drift by applying non-Gaussian degradation $\mathcal{D}(\cdot)$ to historical frames. The training objective for a frame sequence of length $N$ is:
\begin{equation}
\mathcal{L}_{\text{stage2}} = \sum_{i=1}^{N} \lambda_i \mathbb{E}_{\mathbf{z}_{1,i}, \mathbf{z}_{0,i}, t_i} \left| \mathbf{v}_\theta \left( \mathbf{z}_{t,i}, \mathcal{D}_i(\mathbf{z}_{t,<i}^{\text{gt}}), t_i, c_i \right) - (\mathbf{z}_{1,i} - \mathbf{z}_{0,i}) \right|^2_2
\end{equation}
where $t_i$ represents diffusion timesteps, and $\mathcal{D}$ includes multiple degradation. By using non-Gaussian $\mathcal{D}$, we mimic realistic artifacts without corrupting the standard Gaussian denoising process.

\textit{Temporally-progressive objective.} To force the model to handle increasing error accumulation, we design a curriculum where both the task difficulty and supervision intensity increase with the frame index $i$. Specifically, we set:
\begin{equation}
t_i \sim \mathcal{U}(t_{\text{min}}, 1) \cdot \frac{i}{N}, \quad \lambda_i = f(i) \text{ where } \lambda_1 < \lambda_2 < \dots < \lambda_N
\end{equation}
By sampling larger diffusion timesteps $t_i$ and assigning higher loss weights $\lambda_i$ to later turns, we compel the model to develop robust error-correction capabilities, learning to rectify drifts introduced by preceding steps.

\textit{Self-rollout fine-tuning.} Once foundational capabilities are established, we transition from ground-truth conditions $\mathbf{x}_{<i}^{gt}$ to the model's own previous predictions $\tilde{\mathbf{x}}_{<i}$ to bridge the train-inference gap. The fine-tuning loss becomes:
\begin{equation}
\tilde{\mathcal{L}}_{\text{stage2}} = \sum_{i=1}^N \lambda_i \mathbb{E}_{\mathbf{z}_{0,i}, \mathbf{z}_{1,i}, t_i} \left| \mathbf{v}_\theta(\mathbf{z}_{t,i}, \mathcal{D}_i(\text{sg}[\tilde{\mathbf{z}}_{t,<i}]), t_i, c_i) - (\mathbf{z}_{1,i} - \mathbf{z}_{0,i}) \right|^2_2
\end{equation}
where $\text{sg}[\cdot]$ denotes the stop-gradient operation. These generated frames $\tilde{\mathbf{z}}_{<i}$ serve as realistic, imperfect historical conditions. This strategy significantly alleviates the cumulative color drift and identity erosion that often intensify across extended editing sequences.

\paragraph{Stage 3: Distribution Matching Distillation with Causal Consistency Initialization.} To achieve high-efficiency deployment, we employ a two-step distillation pipeline to compress the sampling process into few steps.
Following recent advances in consistency distillation~\cite{luo2023lcm,sauer2023adversarialdiffusiondistillation}, we first initialize model using Causal Consistency Distillation~\cite{zhu2026causal,song2023consistency}, which leverages the frame-level injectivity of the autoregressive teacher to learn a consistent flow map. We optimize the consistency model $G_\theta$ by conditioning on the historical prefix $\mathbf{x}_{gt}^{<i}$ via a causal teacher-forcing objective:
\begin{equation}
\theta^* = \min_\theta \mathbb{E}_{\mathbf{z}_{gt}, \epsilon, t, i} \left[ w(t) d\left(G_\theta(\mathbf{z}_{t,i}, \mathbf{z}^{gt}_{0,<i}, t), G_{\theta^-}(\hat{\mathbf{z}}_{t-\Delta t, i}, \mathbf{x}^{gt}_{<i}, t-\Delta t)\right) \right],
\end{equation}
where $\hat{\mathbf{x}}_{t-\Delta t}^i$ is obtained by solving the ODE from $\mathbf{x}_t^i$ using the Stage 2 teacher model. $d(.,.)$ is a distance under a chosen norm.
To further refine the generation quality and align the student's output with the high-fidelity data distribution, we perform joint optimization using a Distribution Matching Distillation (DMD) objective combined with a flow-matching loss~\cite{yin2024onestepdmd,yin2024improved}. The gradient for the distribution matching is formulated as:
\begin{equation}
\nabla_\theta \mathbb{E}_t [D_{\text{KL}}(p_{\theta,t} || p_{\text{data},t})] = -\mathbb{E}_{\tilde{\mathbf{z}}, t, \tilde{\mathbf{z}}t} \left[ \left(s_{\text{real}}(\tilde{\mathbf{z}}_t, t) - s_{\text{fake}}(\tilde{\mathbf{z}}_t, t)\right) \frac{\partial \tilde{\mathbf{z}}}{\partial \theta} \right],
\end{equation}
where $\tilde{\mathbf{x}}$ is the student-generated image, and $\tilde{\mathbf{x}}_t$ is its noised version. Here, a frozen pretrained model $s_{\text{real}}$ predicts the score under the real data distribution $p_{\text{data},t}$, while an online-trainable model $s_{\text{fake}}$ estimates the score of the current student distribution $p_{\theta,t}$. This hybrid strategy ensures that the model achieves both rapid convergence and superior visual fidelity in few-step multi-turn editing.

\begin{figure*}[t]
	\centering
	\includegraphics[width=0.95\linewidth]{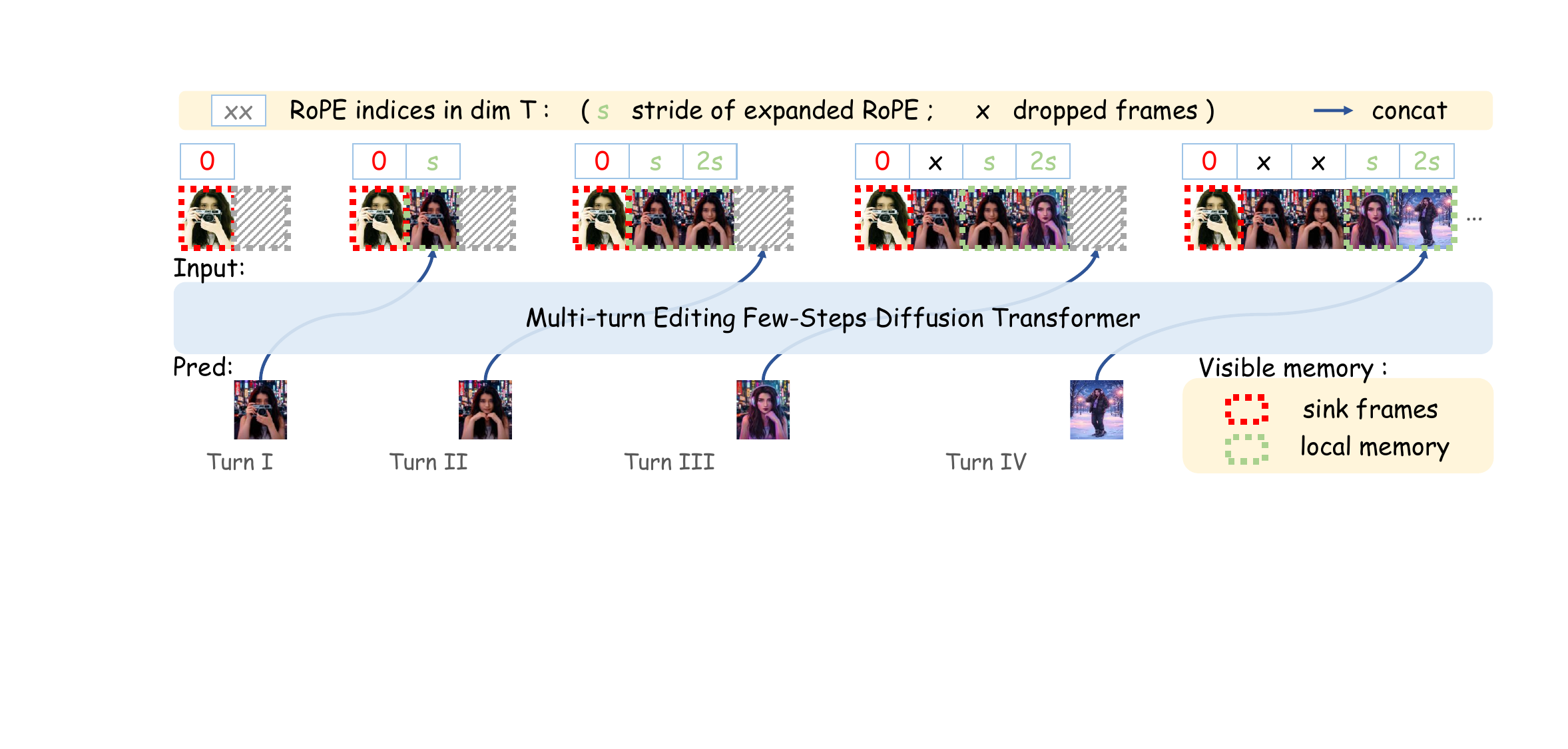}
	\caption{\textbf{Long-chain Streaming Inference Pipeline.} 
(1) \textit{Strided RoPE Indexing}: We anchor the \textbf{sink frame} at index 0, while \textbf{local frames} are assigned indices with a fixed stride $s$. This preserves the "semantic leap" distance learned in Stage 1, preventing identity collapse across long sequences. 
(2) \textit{Streaming Memory Management}: We employ a selective KV cache where the source image serves as a persistent anchor for global consistency, complemented by a sliding window for local coherence. As shown by the 'x' markers, dropped frames do not affect the fixed RoPE indices of the remaining cache, ensuring structural alignment throughout the streaming process.}
	\label{fig:infer}
\end{figure*}

\subsection{Long-chain Streaming Inference Pipeline}
To exploit the extrapolation potential of our causal framework, we implement an efficient inference pipeline featuring a sliding-window KV cache and a persistent attention sink as shown in Fig. \ref{fig:infer}. To support long-turn editing while maintaining the structural priors learned during training, we manage the temporal receptive field and positional embeddings as follows:

\emph{Attention Sink and Memory Management.}
During the $i$-th editing turn, the model performs self-attention over a selective set of historical latents $\Omega_i$ stored in the KV cache. To ensure global subject consistency, we treat the first frame $\mathbf{x}_1$ (the original image) as a persistent anchor that is never evicted from the cache. The rest of the cache follows a sliding-window protocol to maintain local temporal coherence. The set of visible frames $\Omega_i$ is defined as:
\begin{equation}
\Omega_i = {1} \cup {j \mid \max(2, i - W) \leq j < i}
\end{equation}
where $W$ is the window size matching the training sequence length. This mechanism ensures that the model always references the original identity while adapting to the most recent edits.

\emph{Strided RoPE Indexing for Inference Memory.}
We extend the expanded RoPE from Stage 1 into a Strided Positional Indexing scheme for multi-turn inference. As illustrated in Fig. \ref{fig:infer}, instead of assigning sequential indices, we assign RoPE indices based on the frame's role within the visible memory.
Specifically, the Sink frame is anchored at a fixed index $0$ to serve as a permanent structural reference. For the $L$ frames maintained in the Local sliding window, we assign indices using a fixed stride $s$, where $s$ corresponds to the expansion factor used in Stage 1. For a frame at the $j$-th position within the local window ($j \in \{1, \dots, L\}$), its positional index $p_j$ is formulated as:
\begin{equation}
p_j = \begin{cases}
0, & \text{if sink frame} \\
j \cdot s, & \text{if } j\text{-th local frame}
\end{cases}
\end{equation}
This design ensures that even as historical frames are dropped (indicated by 'x' in Fig.~\ref{fig:infer}), the remaining frames in the KV cache retain their relative "semantic distance." By keeping these indices as multiples of $s$, the model consistently perceives a large separation between editing turns, effectively preventing "identity collapse" and ensuring that the multi-turn generation remains within the distribution of the two-frame expanded-distance pretraining.

\begin{figure*}[t]
	\centering
	\includegraphics[width=0.95\linewidth]{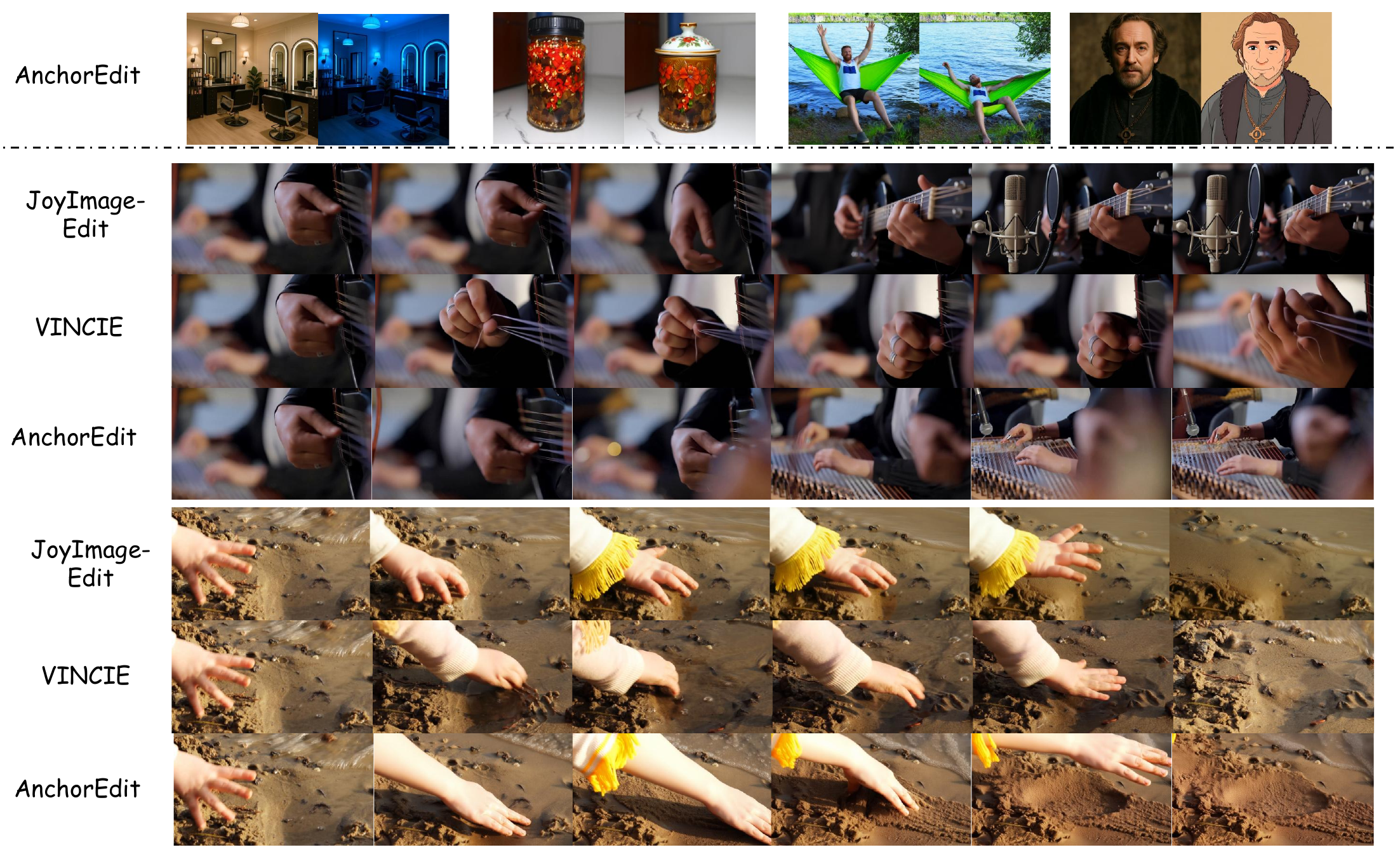}
	\caption{\textbf{Qualitative comparison of image editing results.} 
The top section displays \textbf{single-turn editing} results of our method. 
The bottom section provides a \textbf{multi-turn editing} comparison between our AnchorEdit and existing methods (JoyImage-Edit and VINCIE). 
Our approach demonstrates superior consistency and enhanced semantic alignment across extended editing sequences.} 
	\label{fig:visual}
\end{figure*}

\section{Experiment}
\subsection{Benchmark}
Currently, there is a notable absence of benchmarks for high-resolution, multi-turn editing. To fill this void, we curated a dataset of 50 samples derived from diverse video and image sources. These samples encompass various editing categories and varying numbers of turns, designed to explore the performance ceiling of models in complex multi-step scenarios.
Our evaluation framework is multi-faceted. For each turn, we employ a VLM to assess the results based on two key dimensions: visual consistency and semantic instruction following. Both metrics are scored on a scale of 1 to 10. We implement a termination criterion where an edit is deemed a failure if either score falls below 3, at which point the sequence does not proceed to the subsequent turn. Finally, we report the overall Success Rate, along with the average Consistency Score and Semantic Following Score.

\begin{figure*}[t]
	\centering
	\includegraphics[width=0.95\linewidth]{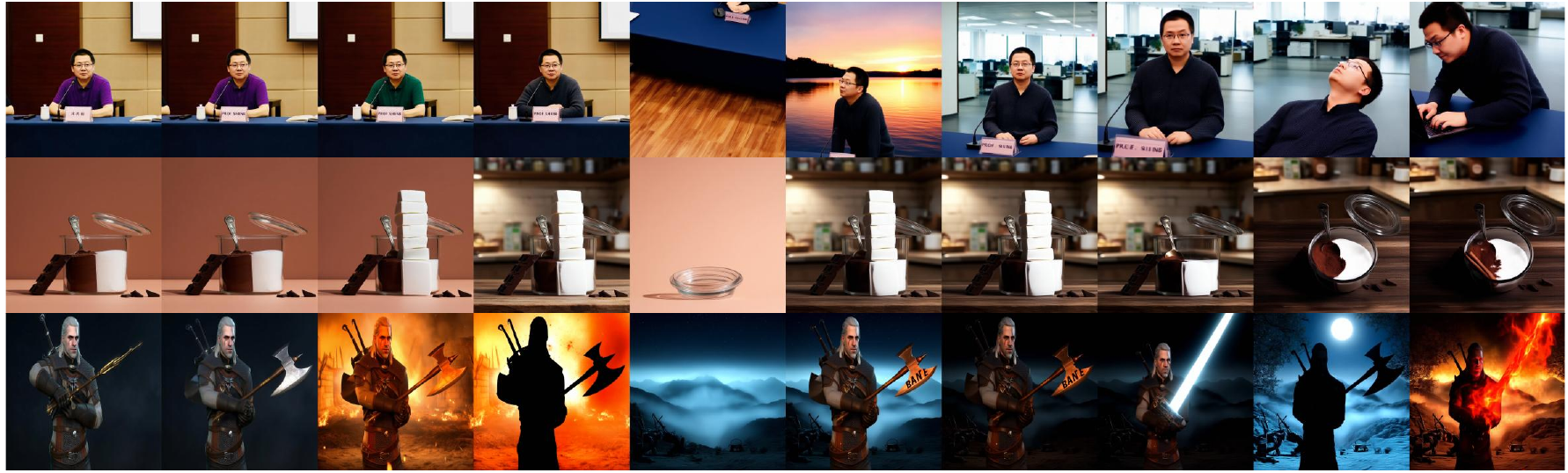}
	\caption{\textbf{Visual results of long-chain 10-turn editing.} 
Each row showcases a sequential 10-turn inference process using our model. }
	\label{fig:10turns}
\end{figure*}

\subsection{Experimental Details}

\paragraph{Data.}
We train our model on two primary datasets: (1) approximately 1 million multi-turn sequences derived from high-motion video clips to ensure temporal stability, and (2) 200,000 diverse image-based editing samples from proprietary generative models to enhance semantic variety. Comprehensive details for data preparation are provided in the Appendix.

\paragraph{Model.} Our framework is built on the WAN2.1-T2V-14B~\cite{wan2025wan} and follows a three-stage training curriculum. It begins with a single-turn stage of 1,500 steps (BS=256) to establish foundational editing capabilities, followed by a multi-turn stage of 1,500 steps (BS=64) to optimize for causal sequential reasoning. Finally, we perform a distillation stage, comprising 500 steps of Consistency Distillation and 1,000 steps of hybrid loss training (both BS=64), to enable few-step inference.

\paragraph{Evaluation.}
We use a 4-step inference strategy with CFG~\cite{ho2022classifier} disabled. For stability, the context window includes the fixed first frame and the 4 most recent edited frames. Performance is evaluated on ImgEdit~\cite{ye2025imgedit} for single-turn tasks, our custom benchmark for multi-turn scenarios, and MSE Bench~\cite{qu2025vincie} for generalizability. Further details and hyperparameters are provided in the Appendix.

\begin{table}[t]
\centering
\caption{\textbf{Results on ImgEdit-Bench}}
\label{tab:imgedit_benchmark}
\begingroup
\setlength{\tabcolsep}{2.2pt}
\renewcommand{\arraystretch}{1.05}
\scriptsize
\resizebox{0.98\textwidth}{!}{
\begin{tabular}{l|ccccccccc|c}
\toprule
\textbf{Model} & \textbf{Add} & \textbf{Adjust} & \textbf{Extract} & \textbf{Replace} & \textbf{Remove} & \textbf{Background} & \textbf{Style} & \textbf{Hybrid} & \textbf{Action} & \textbf{Overall}$\uparrow$ \\
\midrule

ChronoEdit-turbo~\cite{wu2025chronoedit}
& 4.36 & 4.38 & 3.28 & 4.11 & {4.00} & {4.31} & 4.31 & 3.67 & {4.78} & 4.13 \\

JoyAI-Image-Edit~\cite{song2026awakingspatialintelligenceunified}
& 4.63 & 4.52 & {4.32} & 4.71 & {4.76} & 4.53 & 4.88 & 4.09 & 4.69 & 4.57 \\

AnchorEdit
& 4.41 & 4.42 & {3.08} & 4.14 & {3.92} & 4.49 & 4.31 & 3.60 & 4.78 & 4.13 \\

\bottomrule
\end{tabular}
}
\endgroup
\end{table}


\begin{table}[t]
\centering
\caption{Quantitative results on \textbf{Our Multi-turn Benchmark} and \textbf{MSE-Bench}. We report Consistency, Semantic Following, Success Rate for our benchmark, and multi-turn results for MSE-Bench.}
\label{tab:merged_results}
\resizebox{\textwidth}{!}{
\begin{tabular}{l|ccc|ccccc}
\toprule
\multirow{2}{*}{\textbf{Method}} & \multicolumn{3}{c|}{\textbf{Custom Multi-turn Benchmark}} & \multicolumn{5}{c}{\textbf{MSE-Bench (Success Rate per Turn)}} \\
\cmidrule(lr){2-4} \cmidrule(lr){5-9}
 & \textbf{Cons. $\uparrow$} & \textbf{Sem. $\uparrow$} & \textbf{SR (\%) $\uparrow$} & \textbf{Turn-1} & \textbf{Turn-2} & \textbf{Turn-3} & \textbf{Turn-4} & \textbf{Turn-5} \\
\midrule
JoyImage-Edit~\cite{song2026awakingspatialintelligenceunified} & 5.1 & 5.4 & 29.8 & 0.980 & 0.802 & 0.679 & 0.617 & 0.446 \\
VINCIE~\cite{qu2025vincie}        & 5.8 & 5.1 & 34.1 & 0.950 & 0.693 & 0.667 & 0.617 & 0.487 \\
\midrule
\textbf{AnchorEdit} & \textbf{6.2} & \textbf{5.3} & \textbf{52.9} & \textbf{0.950} & \textbf{0.737} & \textbf{0.707} & \textbf{0.623} & \textbf{0.567} \\
\bottomrule
\end{tabular}
}
\end{table}

\subsection{Quantitative Results}
\paragraph{Single-turn Editing.}
On ImgEdit in Tab. \ref{tab:imgedit_benchmark}, our model achieves competitive performance. The model maintains high subject fidelity and instruction following, demonstrating that the single-turn pretraining effectively adapts the 14B backbone while preserving its generative power.

\paragraph{Multi-turn Editing.}
Our model demonstrates superior stability in extended sequences. As shown in Tab. \ref{tab:merged_results}, our model maintains a high Success Rate on both bench. These improvements are primarily driven by our memory mechanism and Self-Rollout fine-tuning, which rectify artifacts and prevent catastrophic drift. Furthermore, our distilled 4-step inference delivers these results with much higher efficiency than full-diffusion baselines.

\subsection{Qualitative Results}
We present qualitative comparisons in Fig. \ref{fig:visual} to demonstrate our model's performance in challenging multi-turn scenarios. A key advantage of our approach is its exceptional visual consistency: while existing baselines often suffer from "identity drift" after only a few turns, our model maintains remarkable subject and background stability even in sequences exceeding ten turns (Fig. \ref{fig:10turns}). These results showcase not only high-fidelity $1024 \times 1024$ generation without over-smoothing but also a robust error-correction capability that prevents recursive artifact amplification.



\begin{table}[ht]
    \centering
    \begin{minipage}{0.49\textwidth}
        \centering
        \caption{Ablation of core components on our benchmark. SR: Success Rate.}
        \label{tab:ablation_core}
        \resizebox{\textwidth}{!}{
            \begin{tabular}{l|ccc}
            \toprule
            \textbf{Configuration} & \textbf{Consist. $\uparrow$} & \textbf{Sem. $\uparrow$} & \textbf{SR (\%) $\uparrow$} \\
            \midrule
            Base & 4.9 & 4.7 & 30.6 \\
            + Expanded RoPE & 5.5 & 5.2 & 45.8 \\
            + Self-rollout (Full) & \textbf{6.2} & \textbf{5.3} & \textbf{52.9} \\
            \bottomrule
            \end{tabular}
        }
    \end{minipage}
    \hfill
    \begin{minipage}{0.47\textwidth}
        \centering
        \caption{Impact of Inference Memory (Sink/Local) on long sequences (\textbf{Turns > 8}).}
        \label{tab:ablation_memory}
        \resizebox{\textwidth}{!}{
            \begin{tabular}{cc|ccc}
            \toprule
            \textbf{Sink} & \textbf{Local} & \textbf{Consist. $\uparrow$} & \textbf{Sem. $\uparrow$} & \textbf{SR (\%) $\uparrow$} \\
            \midrule
            0 & 2 & 4.7 & 5.2 & 33.5 \\
            1 & 2 & 5.3 & 5.6 & 40.3 \\ 
            1 & 4 & \textbf{5.5} & \textbf{5.6} & \textbf{42.7} \\
            \bottomrule
            \end{tabular}
        }
    \end{minipage}
\end{table}
\subsection{Ablation Study}

\paragraph{Core Components.}
We first evaluate the impact of the training pipeline components in Tab. \ref{tab:ablation_core}. Replacing standard sequential positional encoding with \textbf{Expanded RoPE} significantly improves the Consistency Score and Success Rate (SR). By increasing the index distance between frames, the model gains better temporal resolution, which prevents "identity collapse" and the blurring of features across distant turns. The addition of \textbf{Self-rollout Finetuning} provides the most substantial boost, raising the SR to 52.9\%. This confirms that training on model-generated history allows the model to develop self-correction capabilities with preventing minor artifacts during inference.

\paragraph{Inference Memory Mechanism.}
We further analyze the impact of Sink and Local frames on long-chain sequences (Turns > 8) in Tab. \ref{tab:ablation_memory}. Without a visual anchor (Sink=0), the model struggles with identity drift as the sequence progresses. Introducing a \textbf{Sink Frame}—fixing the first frame as a persistent reference—leads to a significant jump in both consistency and SR, effectively anchoring the subject's identity. Increasing the number of \textbf{Local Frames} in the sliding window further enhances short-term coherence. These results demonstrate that our hybrid memory strategy is essential for maintaining stability across extended editing trajectories.

\section{Conclusion}

This paper presents a robust framework for multi-turn image editing by repurposing autoregressive video diffusion models. By introducing a causal training curriculum and a hybrid memory mechanism, our model maintains exceptional subject fidelity and semantic consistency across sequences of 10+ turns. This work demonstrates the potential of causal diffusion architectures in establishing a reliable and scalable foundation for long-horizon interactive visual creation.

{
\small
\bibliographystyle{unsrt}
\bibliography{main}
}
\clearpage
\appendix
\newpage

\maketitle

\section{Experimental Details}
\subsection{Training Data Construction}
To train a robust autoregressive editor capable of high-resolution (1024p) output, we constructed a large-scale, diverse dataset combining natural temporal transitions from videos and complex semantic modifications from synthetic image editing chains.
\paragraph{Video-based Multi-turn Data.}
The primary source for learning subject constancy and motion-aware transitions is a curated collection of approximately 2 million single-shot video clips. These clips were selected based on high motion dynamics and structural complexity to ensure the model learns to handle significant visual changes.
\begin{itemize}
\item \textbf{Curation and Sampling:} From the 2M clips, we sampled 1 million sequences. For each sequence, we extract $N$ frames (where $N$ ranges from 3 to 8) with adaptive temporal spacing. This sampling strategy simulates "editing steps" by treating subsequent frames as the results of sequential instructions.
\item \textbf{Annotation Pipeline:} We utilized a state-of-the-art Vision-Language Model (VLM) to generate dense captions for each frame and, crucially, \textit{delta-instructions} that describe the transformation from frame $t$ to frame $t+1$. This transforms raw video into a "natural editing" dataset where the model learns how objects move and evolve while maintaining strict identity.
\end{itemize}
\paragraph{Synthetic Image-based Editing Data.}
While video data provides excellent identity priors, it lacks the diversity of explicit semantic edits (e.g., "change the texture to gold" or "add a cyberpunk hat"). To fill this gap, we generated a specialized dataset of 200,000 multi-turn editing sequences.
\begin{itemize}
\item \textbf{Synthesizing Edit Chains:} We employed a "Teacher-Critic" framework. A Large Language Model (LLM) first generates a logically consistent 5-step editing plan for a source image. A high-quality closed-source diffusion editor then executes these steps.
\item \textbf{Diversity:} This subset focuses on diverse categories including global style transfers, local object manipulation, and attribute editing. Although smaller in scale than the video data, its high semantic variance prevents the model from overfitting to the simple physical transitions found in natural videos.
\end{itemize}
\paragraph{Quality Control and Preprocessing.}
To ensure the high-fidelity performance of AnchorEdit, we applied a multi-stage filtering pipeline:
\begin{itemize}
\item \textbf{Resolution and Aesthetics:} All data were filtered to ensure a minimum resolution of 1024p. We used an internal aesthetic scorer to remove samples with motion blur, watermarks, or low-quality compression artifacts.
\item \textbf{Alignment Filtering:} We calculated CLIP-based text-image alignment scores for every turn. Sequences where the visual change did not match the generated instruction were discarded.
\item \textbf{Identity Consistency:} For the synthetic image data, we applied a face-reid and DINO-v2 feature similarity check across the editing chain. Only sequences that maintained a high similarity score (ensuring the subject did not "morph" into a different entity) were retained for Stage 2 training.
\end{itemize}
This hybrid dataset ensures that AnchorEdit is both a stable "anchor" for subject identity (via video priors) and a versatile "editor" for complex instructions (via synthetic image data).

\subsection{Model Training Details}
We initialize AnchorEdit using the \textbf{Wan2.1-T2V-14B} backbone, a state-of-the-art autoregressive video generation model. The training is divided into three specialized stages to transition the model from a video generator to a robust, high-resolution causal editor.
\paragraph{Stage 1: Multi-Resolution Single-Turn Pretraining.}
In the first stage, we establish foundational editing capabilities and identity preservation.
\begin{itemize}
\item \textbf{Dataset:} We utilize a massive dataset of 3 million samples, including public single-turn editing datasets and our internally constructed image-editing chains.
\item \textbf{Multi-Resolution Strategy:} To ensure the model is resolution-agnostic up to 1024p, we employ a multi-resolution training strategy where the total pixel count is $\leq 1024 \times 1024$, with aspect ratios ranging up to 4:1. This ensures the model can handle various canvas shapes (e.g., cinematic wide or mobile portrait).
\item \textbf{Optimization:} We train for 1,500 steps with a batch size of 256. We use a fixed learning rate of $1e-5$, with AdamW optimizer parameters set to $\beta_1=0.9, \beta_2=0.999$, and a weight decay of $0.01$. The maximum gradient norm is clipped at $10.0$ to ensure stability.
\end{itemize}
\paragraph{Stage 2: Causal Multi-Turn and Self-Rollout Training.}
This stage optimizes the model for sequential reasoning and robustness against error accumulation. We split this stage into two progressive phases:
\begin{itemize}
\item \textbf{Warm-up Phase (Steps 0--500):} We train exclusively on ground-truth (GT) history frames to establish basic causal reasoning. The self-rollout probability is set to 0.
\item \textbf{Self-Rollout Phase (Steps 500--1,500):} To mitigate exposure bias, we introduce a \textbf{Self-Rollout} strategy. For each sequence, internal frames (excluding the first and last) have a 0.3 probability of being replaced by the model’s own previous-step predictions. While these rollout frames receive noise and degradation like GT frames, we apply a \textit{loss mask} to them, ensuring gradients are only computed for the target frame.
\item \textbf{Robustness via Degradation:} To simulate the distribution shift encountered during inference, we apply three types of degradations to historical frames with a 0.3 trigger probability each: (1) \textit{Spatial Blur} (excluding Gaussian blur to avoid interference with the diffusion process), (2) \textit{JPEG Compression} artifacts, and (3) \textit{Color Shifting} via channel perturbations. Furthermore, we pass frames through the VAE encoder/decoder loop to simulate latent-space distribution shifts.
\item \textbf{Temporal Scheduling:} We increase the difficulty of the task for later turns in a sequence. Specifically, the noise intensity, degradation strength, and loss weights are scaled progressively. The loss weight is set to $1.0$ for the second turn and increases linearly to $2.0$ for the final turn.
\end{itemize}
\paragraph{Stage 3: Causal Consistency Distillation.}
The final stage focuses on accelerating inference while maintaining the causal properties developed in Stage 2.
\begin{itemize}
\item \textbf{Consistency Distillation (CD):} We adapt the consistency distillation objective to our causal framework. The training setup mirrors the "Causal Forcing" strategy from Stage 2.
\item \textbf{Procedure:} We extract the initial image and the full sequence of prompts from our training set. The model is trained to map any point on the ODE trajectory to the solution in a single (or few) steps. This phase comprises 1,500 steps (500 for initial CD and 1,000 for few-step refinement) with a batch size of 64, resulting in a high-efficiency editor capable of generating high-quality results in just 4 sampling steps.
\end{itemize}

\subsection{Benchmark}
\begin{tcolorbox}[
    colback=white, 
    colframe=black!75, 
    title=\textbf{System Prompt for Multi-turn Evaluation},
    fonttitle=\bfseries,
    arc=4pt,
    outer arc=4pt,
    boxrule=0.8pt,
    left=10pt,
    right=10pt,
    top=10pt,
    bottom=10pt
]
\small
\textbf{Role:} You are an expert visual quality inspector specialized in multi-turn image editing. Your task is to evaluate the quality of an edit based on the initial state, the previous state, and the current editing instruction.

\textbf{Input Information:}
\begin{enumerate}
    \item \textit{Initial Image ($I_0$):} The original image before any edits (The identity anchor).
    \item \textit{Previous Image ($I_{t-1}$):} The image from the last turn.
    \item \textit{Current Image ($I_t$):} The result of the current edit.
    \item \textit{Instruction:} The specific modification requested for this turn.
\end{enumerate}

\textbf{Task:} Assign two scores (1-10) based on the following criteria:

\textbf{1. Visual Consistency Score (CS):}
Compare $I_t$ with $I_0$ and $I_{t-1}$. 
\begin{itemize}
    \item \textbf{9-10:} Subject identity, background structure, and non-edited details are perfectly preserved.
    \item \textbf{7-8:} Subject is clearly the same, but very minor flickering or texture changes occur.
    \item \textbf{5-6:} Subject is identifiable, but shows noticeable drift (e.g., facial feature shifts, clothing pattern changes).
    \item \textbf{3-4:} Significant identity drift; subject looks like a "different person/object" of the same category.
    \item \textbf{1-2:} Identity collapse or image corruption.
\end{itemize}

\textbf{2. Semantic Following Score (SF):}
Compare $I_t$ with $I_{t-1}$ and the Instruction.
\begin{itemize}
    \item \textbf{9-10:} Instruction is followed precisely with high visual quality and natural blending.
    \item \textbf{7-8:} Modification is clear and correct, with only minor artifacts or slight inaccuracies in extent.
    \item \textbf{5-6:} Instruction is partially followed but lacks precision (e.g., wrong color shade, object added unrealistically).
    \item \textbf{3-4:} The edit is attempted but the result is barely recognizable or visually garbled.
    \item \textbf{1-2:} The instruction was ignored, or the edit is irrelevant.
\end{itemize}

\textbf{Output Format:}
Return only a JSON object: \texttt{\{"Consistency\_Score": X, "Semantic\_Score": Y, "Reasoning": "..."\}}
\end{tcolorbox}

To rigorously evaluate AnchorEdit, we move beyond simple CLIP-based metrics, which often fail to capture fine-grained identity drift in multi-turn scenarios. Instead, we utilize a GPT-4o-based evaluation protocol that simulates a human user's judgment across two critical dimensions:

\emph{Identity Anchoring vs. Instruction Adherence.}
A common failure mode in multi-turn editing is "over-editing," where the model follows the instruction but completely loses the original subject's identity. By providing the VLM with the Initial Image ($I_0$) as a global reference and the Previous Image ($I_{t-1}$) as a local reference, the Consistency Score (CS) specifically penalizes the "identity leakage" that accumulates over time.

\emph{Sequential Failure and Chain Termination.}
In our benchmark, we emphasize the \textit{sequential stability}. Unlike single-turn benchmarks, the error in turn $n$ directly affects turn $n+1$. Our scoring system incorporates a "Success Threshold" (Score $\ge 3$). If a model produces a result that is either visually corrupted (CS < 3) or completely ignores the prompt (SF < 3), the editing chain is considered "broken." This reflects real-world usability: once a user's subject is lost, subsequent edits are meaningless.

\emph{High-Resolution Sensitivity.}
Since our benchmark is conducted at 1024p resolution, the VLM is instructed to look for high-frequency artifacts (e.g., JPEG noise, texture blurring) in its reasoning process. The CS score specifically looks for background stability—ensuring that an edit to a person's hat does not inadvertently change the texture of the trees behind them.

\section{More Ablations}
\begin{table}[h]
\centering
\caption{\textbf{Ablation study on Reweighted FLow Matching Loss.} We compare the performance of the model trained with a fixed loss weight versus our proposed linear weighting strategy ($1.0 \to 2.0$ from the second to the final turn).}
\label{tab:ablation_loss_weight}
\begin{tabular}{@{}lccc@{}}
\toprule
\textbf{Configuration} & \textbf{CS $\uparrow$} & \textbf{SF $\uparrow$} & \textbf{SR $\uparrow$} \\ \midrule
Fixed Loss Weight ($w_t = 1.0$) &6.0 &5.0 &.50.1 \\
Temporal Loss Weighting ($1.0 \to 2.0$) & \textbf{6.2} & \textbf{5.3} & \textbf{52.9} \\ \bottomrule
\end{tabular}
\end{table}
\paragraph{Reweighted FLow Matching Loss}
A core challenge in autoregressive multi-turn editing is the "exposure bias" and the resulting error accumulation. As the number of editing turns increases, the input frames for the current turn inevitably contain more artifacts or subtle identity drifts from previous steps. To address this, we proposed a Temporal Loss Weighting strategy during the second stage of training.
These results validate that temporal scheduling is essential for mastering long-horizon editing tasks, as it effectively guides the model to prioritize robustness as the difficulty of the task increases across turns.

\begin{table}[h]
\centering
\caption{Ablation study on degradation strategies for historical frames. We compare the effects of no degradation, fixed-intensity degradation, and our proposed progressive degradation across 10-turn editing sequences.}
\label{tab:ablation_degradation}
\begin{tabular}{@{}lccc@{}}
\toprule
\textbf{Configuration} & \textbf{CS $\uparrow$} & \textbf{SF $\uparrow$} & \textbf{SR $\uparrow$} \\ \midrule
    Base (Clean History) & 4.9 & 4.7 & 30.6  \\
    Fixed Degradation & 6.2 & 5.2 & 49.7 \\
    Progressive Degradation (Full) & \textbf{6.2} & \textbf{5.3} & \textbf{52.9} \\
    \bottomrule
    \end{tabular}
\end{table}

\paragraph{The Necessity of Progressive Degradation.}
In an autoregressive inference chain, the quality of historical frames is not constant; rather, it exhibits a "decaying" trend where late-turn frames accumulate more artifacts than early-turn ones. We conduct an ablation study to verify whether the training noise should reflect this temporal distribution.
\begin{itemize}
\item \textbf{Fixed Degradation:} In this setting, we inject a constant level of noise, blur, and compression to all historical frames regardless of their turn index. As shown in Table \ref{tab:ablation_degradation}, this improves the Success Rate (SR) from 30\% to 49.7\% compared to the clean-history baseline. This suggests that simply exposing the model to imperfect inputs helps mitigate exposure bias to some extent.
\item \textbf{Progressive Degradation (Full):} Our proposed strategy scales the intensity of degradations (e.g., Gaussian noise $\sigma$ and VAE-loop blur) linearly with the turn index $t$. This configuration yields the best performance across all metrics (SR: 52.9\%, Consist.: 6.2).
The reason for this improvement is twofold:
(1) \textbf{Distribution Alignment:} Progressive degradation accurately simulates the inference-time environment where the "anchor" frames in later turns are significantly noisier.
(2) \textbf{Curriculum Learning:} It acts as a form of curriculum learning, where the model first learns to preserve identity from relatively clean history in early turns, and gradually learns more robust "rectification" skills to handle heavily corrupted history in later turns.
The gap between Fixed and Progressive degradation (3.2\% in SR) demonstrates that teaching the model to handle "increasingly difficult" history is key to maintaining a stable and consistent 10-turn editing chain.
\end{itemize}

\section{More Visual Results}

\subsection{Training Data Visualization}
\begin{figure*}[t]
	\centering
	\includegraphics[width=0.95\linewidth]{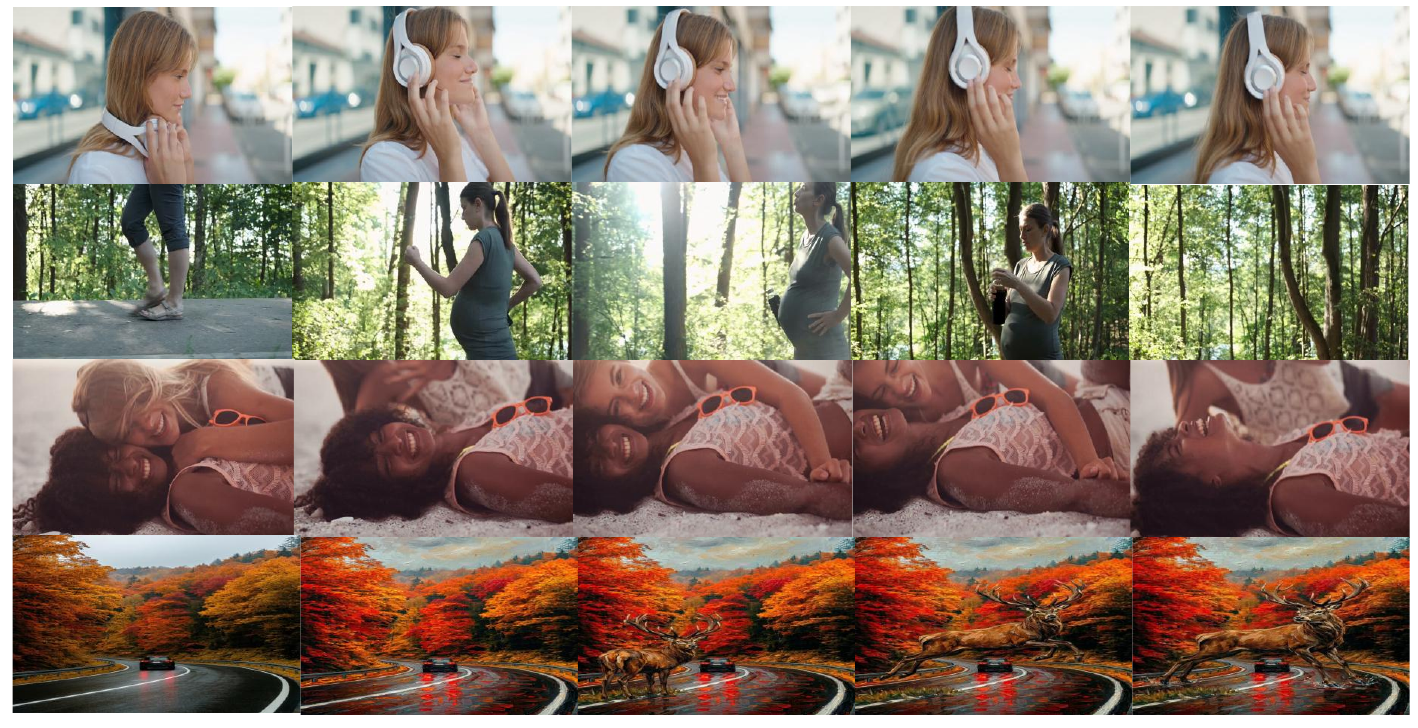}
	\caption{\textbf{Visualization of training data.} } 
	\label{fig:training}
\end{figure*}

In Fig.~\ref{fig:training}, we provide a comprehensive visualization of the synthesized training samples used in our two-stage training pipeline. Unlike single-turn editing datasets, our training data is specifically designed to capture the complex dynamics of multi-turn interactions. Each sample consists of an initial "anchor" image followed by a sequence of logically connected editing instructions and their corresponding outcomes. 

As illustrated, these samples cover a diverse range of editing scenarios, including local object manipulation (e.g., changing clothes, adding accessories), global style transfer, and background modifications. Our model is required to learn not only the specific semantic "delta" defined by each new instruction but also the strict preservation of identity and spatial consistency across the entire chain. By visualizing the training pairs, we demonstrate how the model is exposed to varying degrees of "historical noise" and "simulated degradations" (as discussed in the ablation study), which is crucial for mastering the transition from one editing state to the next without losing the core features of the subject.

\subsection{Benchmark Visualization}
\begin{figure*}[t]
	\centering
	\includegraphics[width=0.95\linewidth]{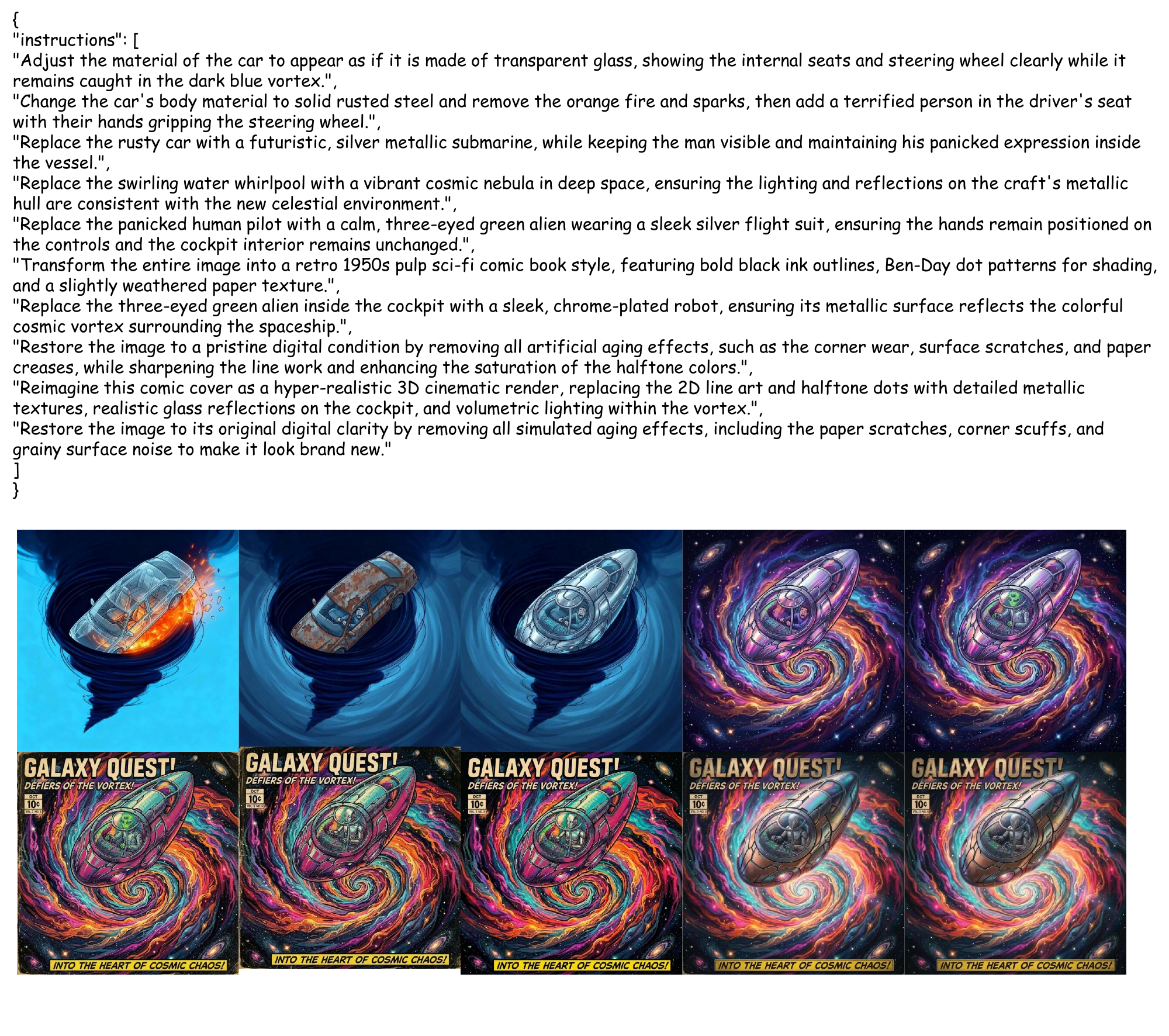}
	\caption{\textbf{Visual results of our models on our proposed benchmark.} } 
	\label{fig:s1}
\end{figure*}

\begin{figure*}[t]
	\centering
	\includegraphics[width=0.95\linewidth]{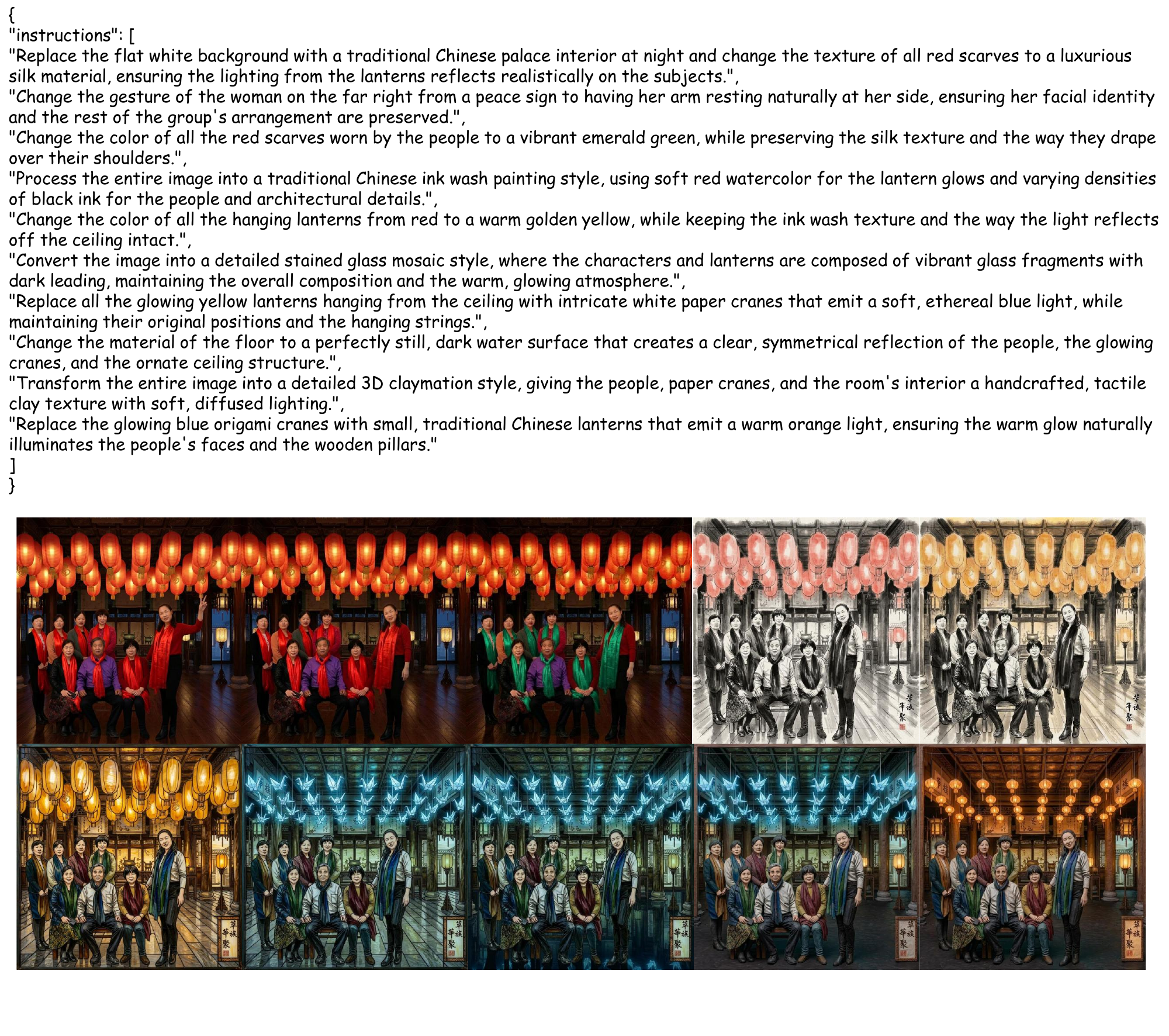}
	\caption{\textbf{Visual results of our models on our proposed benchmark.} } 
	\label{fig:s2}
\end{figure*}

\begin{figure*}[t]
	\centering
	\includegraphics[width=0.95\linewidth]{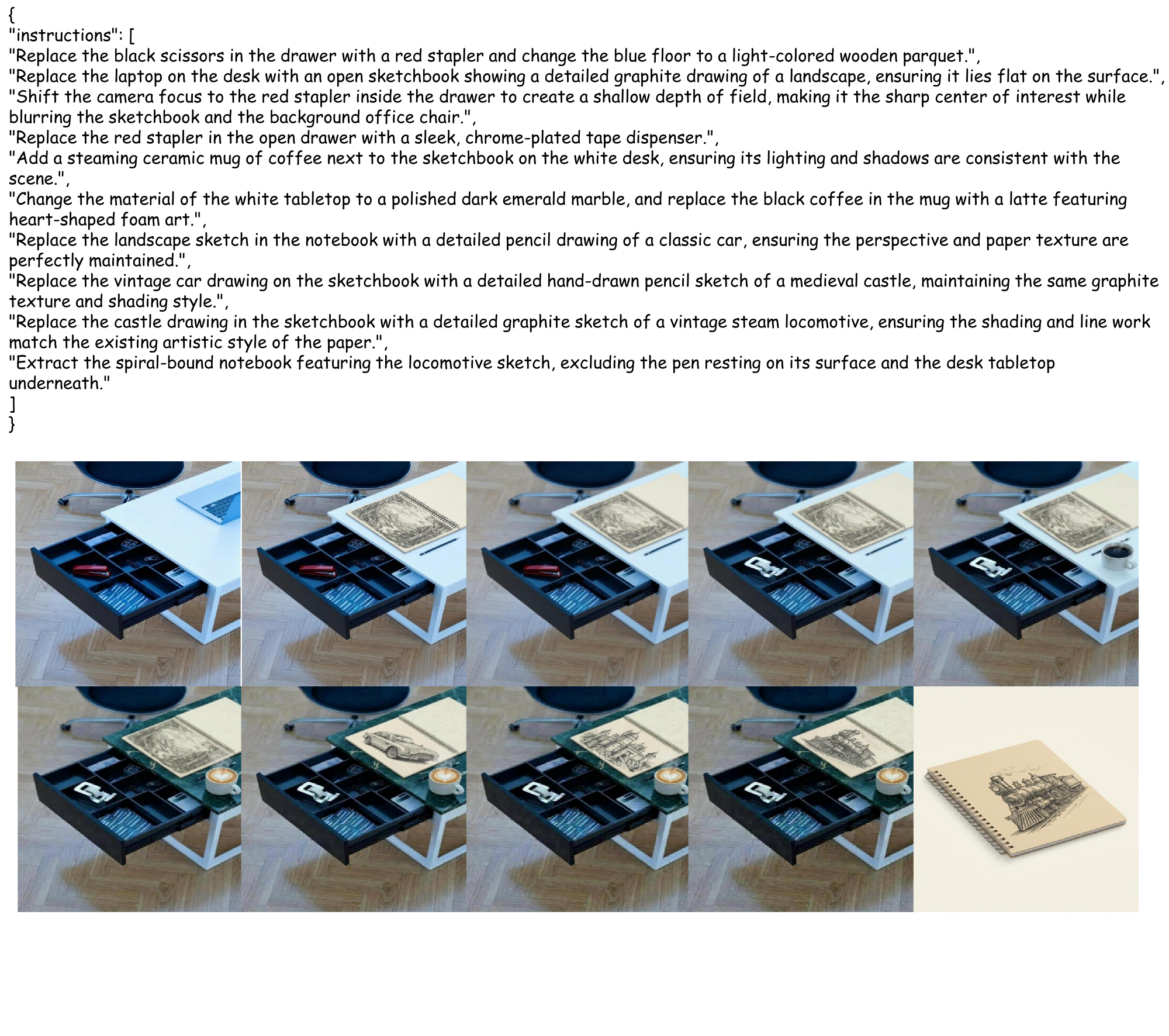}
	\caption{\textbf{Visual results of our models on our proposed benchmark.} } 
	\label{fig:s3}
\end{figure*}

To further validate the robustness and generalization of AnchorEdit, we present extended visual results on our multi-turn benchmark in Fig.~\ref{fig:s1}, Fig.~\ref{fig:s2}, and Fig.~\ref{fig:s3}. These sequences represent some of the most challenging cases in our evaluation set, involving up to 10 consecutive editing turns with diverse and sometimes conflicting instructions.

A key observation from these results is that, despite the extreme sequence length, our model maintains excellent temporal-consistency and high-fidelity textures. Specifically:
\begin{itemize}
    \item \textbf{Identity Preservation:} Even in the 10th turn, the primary subject's fine-grained features (e.g., facial structure, garment textures) remain consistent with the original frame, effectively avoiding the "identity drift" common in standard autoregressive models.
    \item \textbf{Semantic Precision:} The model demonstrates a high degree of instruction-following capability, accurately executing complex commands (e.g., "now change the background to a snowy mountain while keeping the red jacket") without being confused by the long history of previous edits.
    \item \textbf{Quality Stability:} There is no significant accumulation of VAE-related artifacts or blurring, proving that our progressive degradation training strategy effectively teaches the model to "rectify" potential errors at each step.
\end{itemize}
These visualizations confirm that AnchorEdit successfully bridges the gap between short-term image editing and long-horizon, conversation-driven content creation.

\section{Limitation}
Despite its robustness, our framework has a few limitations. First, the autoregressive nature of the model results in inference time that scales linearly with the number of editing turns, which may limit its use in real-time interactive scenarios. Second, while our memory mechanism excels at maintaining subject identity, it may occasionally lead to "over-consistency," making it challenging to perform radical semantic transformations where the original subject's structure needs to be fundamentally altered. Future work will explore more efficient inference techniques and adaptive mechanisms to balance preservation with flexibility.

\end{document}